# Imitate then Transcend: Multi-Agent Optimal Execution with Dual-Window Denoise PPO

**Jin Fang** [*]
University of California Berkeley
Berkeley, CA 94720
`jin_fang@berkeley.edu`

**Jiacheng Weng** [*]
University of Waterloo
Waterloo, ON, CA N2L3G1
`jiacheng.weng@uwaterloo.ca`

**Yi Xiang** [†]
University of Toronto
Toronto, ON, CA M5S1A1
`s.xiang@mail.utoronto.ca`

**Xinwen Zhang** [†]
Columbia University
New York, NY 10027
`xz2782@columbia.edu`

## Abstract

A novel framework for solving the optimal execution and placement problems using reinforcement learning (RL) with imitation was proposed. The RL agents trained from the proposed framework consistently outperformed the industry benchmark time-weighted average price (TWAP) strategy in execution cost and showed great generalization across out-of-sample trading dates and tickers. The impressive performance was achieved from three aspects. First, our RL network architecture called Dual-window Denoise PPO enabled efficient learning in a noisy market environment. Second, a reward scheme with imitation learning was designed, and a comprehensive set of market features was studied. Third, our flexible action formulation allowed the RL agent to tackle optimal execution and placement collectively resulting in better performance than solving individual problems separately. The RL agent's performance was evaluated in our multi-agent realistic historical limit order book simulator in which price impact was accurately assessed. In addition, ablation studies were also performed, confirming the superiority of our framework.

## 1 Introduction

Order execution strategies are basic yet vital elements for trading systems, because they impact the profitability of high-level investment decisions and trading strategies, especially those involving large order volumes. Common execution strategies usually divide a large order into multiple sub-orders and get them filled separately [1]. To minimize the execution cost, i.e., the total cost of buying or selling a certain amount of securities within a predefined time period, the sub-order timing, quantity, and price (placement) are widely studied by both industry professionals and academic researchers [1, 2, 3]. The optimization of these three parameters leads to two well-known problems: optimal execution (sub-order timing and quantity optimization) and optimal placement (sub-order pricing optimization) [4].

Development of the optimal execution and placement strategies often relies on market simulation. However, complex and unknown market dynamics pose significant challenges to the validation of these strategies. Early studies often addressed the optimal execution and placement separately under the framework of optimal control with various unrealistic assumptions. For example, the optimal execution was first studied by Bertsimas and Lo [2]. They leveraged stochastic dynamic programming to derive explicit closed-form solutions and also calculated the expected cost of the best execution strategy. However, their random walk price path and naive price impact functions ignored the demand and supply dynamics in the real financial market. Subsequent work focused on bringing Bertsimas and Lo's framework closer to the realistic scenarios. Almgren and Chriss [5] incorporated the execution risk into the formulation and obtained closed-form solutions of optimal execution for specified risk tolerance levels. Obizhaeva and Wang [6] further added the intertemporal limit order book (LOB) model to capture supply and demand in the market. However, their assumptions about the shape of LOB dynamics were still not realistic, such as the static density of limit orders at each price level and the linear relationship between mid-price shift and the total trade size. For the optimal placement problem, Guo *et al.* [7] derived an optimal placement strategy under a correlated random walk price model. However, the interaction between execution and placement strategies has not been studied.

---

[*]Both authors contributed equally to this research.
[†]Both authors contributed equally to this research.

The way researchers tackled the optimal execution and placement problems changed after the high-frequency order message data from NASDAQ have become available. Nevmyvaka *et al.* [8] carried out the first optimal execution study based on historical LOB simulation and showed that reinforcement learning (RL) can reduce the execution cost compared to Time-weighted Average Price (TWAP) strategy. However, the RL agent's actions have no impact on LOB and future environment observations. More recently, several researchers attempted optimal execution problems leveraging more advanced RL algorithms, such as deep Q-learning [9], Double Deep Q-Network (DDQN) [10], and Proximal Policy Optimization (PPO) [11, 12]. Ning *et al.* [13] designed a market order-based formulation and trained an agent using DDQN, which outperformed the standard TWAP benchmark. However, the associated execution prices were assumed to be mid-price at the time of order placement. Dabérius *et al.* [14] studied the problem using both value-based (DDQN) and policy-based (PPO) RL approaches under stochastic price path simulations. They showed that RL algorithms can reach theoretical optimal strategy solved by dynamic programming and can generate dynamic strategies to beat TWAP. Nevertheless, neither the benchmark strategy nor the agent's activities in these experiments had an accurate impact on the LOB and therefore led to unrealistic execution cost and performance evaluations.

More recent studies have used realistic market simulations allowing interactions between the replayed historical LOB and the agents. Fang and Karpe *et al.* [15] set up a multi-agent simulation environment to study optimal execution and showed that DDQN agent can converge to TWAP strategy. Schnaubelt *et al.* [16] used similar LOB replay techniques on crypto-currencies and focused on optimal placement while assuming fixed execution strategies. As optimal execution and optimal placement are closely related, it is important to study two problems collectively so that the obtained execution strategies are more complete, systematic, and potentially more effective in execution cost minimization than those to deal with each problem alone.

## 1.1 Contributions

The goal of this study is to investigate RL-based optimal strategies for solving both optimal execution and placement problems in a realistic and interactive market simulator. The main contributions of this work are listed below:

1. We developed a computationally efficient and Gym-compatible multi-agent discrete event simulator based on an existing framework [17], which could precisely reconstruct the LOB using nanosecond-level historical message data, in which multiple agents with different objectives and trading strategies can interact with each other. [3]

2. We designed an innovative RL problem formulation with a rich feature set, flexible agent actions, and a novel reward formulation based on imitating and competing with other agents in the simulation.

3. We constructed a novel RL network architecture called Dual-Window Denoise PPO for efficient temporal learning in the noisy financial market environment, which showed superior performance over the original PPO formulation with fully connected networks.

---

[3] We intend to make the simulator open-source in the future to accelerate the academic research on similar topics.

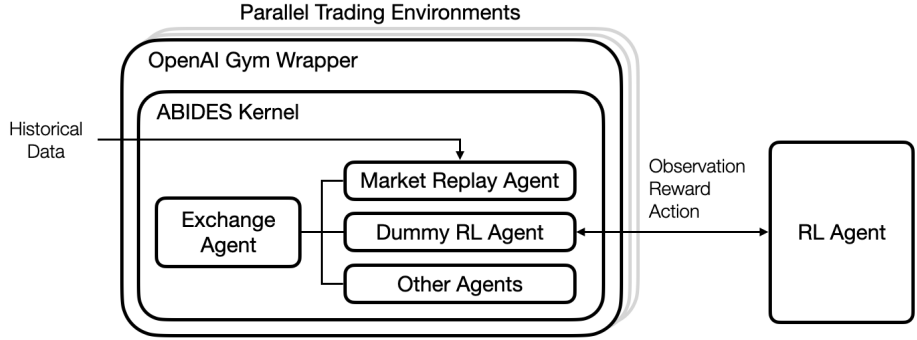

Figure 1: The architecture of the trading environment consists of interactions among internal and external agents. The dummy RL agent, which is an execution individual, observes the market and places orders by communicating with the external RL agent; The inter-agent communication is managed by the ABIDES kernel, which is formatted into the OpenAI Gym style for easy docking and environment parallelization with existing RL packages.

4. We produced RL-based strategies and evaluated them in the realistic market simulation that we built. Our strategies not only beat the industry benchmark TWAP strategy, but also demonstrated strong out-of-sample and cross-ticker generalization ability.

To our best knowledge, this is the first research that trained deep RL agents with high flexibility on both order placement and execution, directly competing and beating the benchmark under the realistic multi-agent LOB simulation, in which execution cost and price impact were correctly calculated.

## 2 Environment Formulation

An environment is a simulated playground for the RL agent to collect experience. It provides the RL agent with observations (current states of the environment) and reward signals (scores of the agent's intermediate performance), and takes the agent's action to progress its states. The backbone of the environment that simulates the trading market is an agent-based simulation platform called Agent-Based Interactive Discrete Event Simulator (ABIDES) [17]. The original ABIDES infrastructure is incompatible with the existing OpenAI Gym environment style for two reasons. First, it has a centralized kernel managing all inter-agent communications in the form of messages, thus, it requires the RL agent to be defined within the kernel. Second, the kernel does not have step control flow and can only stop when the message queue is empty, preventing synchronization between the environment and RL algorithms. To overcome these barriers, we reformulated the kernel architecture by wrapping the centralized kernel into the OpenAI Gym convention as shown in Fig. 1 which enabled precise synchronization between the ABIDES kernel and the Gym functionality. To allow direct information pass-through between the trading environment and the external RL algorithms, a Dummy RL Agent is introduced to manage order placement and observation/reward calculation.

During LOB simulations in Fig. 1, the exchange agent matches the incoming orders, sends specific order status updates, and broadcasts the latest market status to agents who requested subscriptions. The available market information includes the raw LOB states and aggregated candlestick style data of a predefined frequency. A market replay agent sends the historical bid and ask message data to the exchange agent. Other additional agents (e.g., Dummy RL Agent and TWAP Execution agent) can also send orders to the exchange and receive corresponding order status feedback. Each day is considered one episode, and the dummy RL agent is allowed to observe the market and place orders every minute. At the end of each one-minute interval, the unexecuted portion of the orders will be cancelled to avoid confusion in reward assignment (i.e., the temporal mismatch between the agent's action and the delayed reward signals).

Table 1: Overview of all features included as the RL agent's observation.

| Categories | Feature Names |
| --- | --- |
| Task-specific | Remaining time<br>Remaining quantity<br>Order fulfillment percentage |
| Microstructure | Bid-ask spread<br>Direction of trades<br>Effective spread<br>Price improvement<br>Smart price<br>Bid-ask volume imbalance<br>Log Return<br>Mid-price Volatility |
| Technical analysis | Williams R Indicator<br>Exponential Moving Average (EMA)<br>Know Sure Thing (KST) Indicator<br>Moving Average Convergence Divergence (MACD)<br>Keltner Channel<br>Donchian Channel<br>Bollinger Bands<br>Ichimoku Cloud |
| LOB | Top 5-level bid/ask prices<br>Top 5-level bid/ask volumes |

## 2.1 Features

In this work, we provide a comprehensive list of features as the execution agents' observation including task-specific features, microstructure features, technical analysis features, and raw LOB information as summarized in Table 1.

Three task-specific features describe the macro and local states of the execution task. Remaining time and remaining quantity provide essential information about the overall progress of the execution task. Their values are normalized using the following equations: $\hat{t} = \frac{T-2t}{T}, \hat{V}_t^r = \frac{2V_t^r - V_0^r}{V_0^r}$, where $t, V_t^r$ are the current time and remaining quantity, respectively, $T, V_0^r$ are the total time and total quantity, respectively. Both $\hat{t}$ and $\hat{V}_t^r$ are normalized to a range of $[-1, 1]$ for efficient RL agent learning. Order fulfillment percentage captures local execution states by measuring the ratio between the executed quantity and the submitted quantity of the latest order from the agent.

We constructed microstructure features based on classical financial economic literature [18, 19, 20, 21, 22, 23, 24, 25, 26, 27], covering the market's liquidity, short term price movement, and order flows in the LOB. These features are computed directly from the dynamic LOB states broadcast from the exchange agent, which correctly reflects the order impacts of all market participants in the simulation, including the RL agent.

The market liquidity is measured in bid-ask spread [23, 24] as well as the transaction cost related variables including direction of trade, effective spread and price improvement. Bid-ask spread is the difference between top-level bid price $P^b$ and ask price $P^a$ calculated as $S_t = P_t^a - P_t^b$. Direction of trade $D_t$ is based on Lee-Ready algorithm ([18]), indicating the liquidity providers in the market (i.e., whether the last observed trade is buyer or seller initiated). Effective spread [25, 19, 26, 27] is calculated as $ES_t = 2D_t(P_t^e - P_t^m)/P_t^m$, which is the difference between the execution price $P^e$ and the mid-price $P^m$ (i.e., midpoint of the bid and ask prices) at the time of transaction $t$, where $D_t$ is the direction of trade. Price improvement calculates the amount of execution price improvement with respect to the quote, as a result of dark trade or execution against a hidden order ([28]). It is the difference between the national best offer (NBO) and the execution price for a marketable buy order, or the difference between the national best bid (NBB) and the execution price for a marketable sell order.

Four other features, including smart price, bid-ask volume imbalance, log return, and mid-price volatility, reflect the price discovery process (i.e. the process in which information are incorporated into securities price through trading) in the market [29]. Smart price [20] is a inverse volume adjusted price calculated as $(P_t^a V_t^b + P_t^b V_t^a)/(V_t^a + V_t^b)$. Bid-ask volume imbalance, which represents the driving force of the stock price [21], is calculated as $(V_t^b - V_t^a)/(V_t^a + V_t^b)$ [22], where $V^b, V^a$ are the bid and ask volumes, respectively. Log return $log(P_t/P_0)$ and mid-price volatility $std(log(P_t^m/P_0^m))$ [30, 23] reflect the scale and dispersion of change in stock price, where $P, P^m$ are the last transacted price and mid-price, respectively, $std$ calculates the standard deviation (SD). In addition, appropriate transformations (e.g., tanh operation) are applied on the microstructure features to constrain the value range within [-1, 1].

While microstructure features focus on extracting granular information from LOB structure details and the determinants of spreads and quotes, technical analysis indicators are usually used to identify trading signals and price trends. We picked eight technical features to capture the higher-level market price momentum and trends as the supplement to the previously mentioned feature set[31, 32]. Williams R is an indicator of the price movement momentum, and it measures the overbought/oversold levels in the market. EMA, KST and MACD are famous indicators that track the price trend using multiple signal lines (signal line crossover indicates new trends). Keltner Channel, Donchian Channel, Bollinger Bands, Ichimoku Cloud are more sophisticated technical indicators. They calculate different bands based on high/low prices and moving averages of mid prices to indicate support and resistance levels in the market. All technical features are normalized into [-1, 1] as well.

As the RL agent requires temporal observation (explained in section 3), all features are stored in a memory buffer of the dummy RL agent at the end of each step. The memory buffer disposes old feature records that are no longer needed as the simulation progresses, in order to improve the memory efficiency when multiple environment instances are launched for parallel agent learning.

### 2.2 Action Formulation

Given market observations from the environment, the RL agent generates actions that determine the order volumes. In this work, the agent's action is implemented using the idea of policy blending (i.e., using a learning-based policy to modify or modulate an existing baseline policy). This idea has been widely used in learning-based robotic control research for complex systems to achieve high learning efficiency and control robustness [33, 34]. In this work, we use the TWAP-like strategy as the baseline policy which places orders with the same volumes at the level-one price of the LOB with a fixed time interval. Then, the agent actions, which represent the scaling factors across top-three LOB levels, modify the TWAP strategy given market observations as shown in the following equation:

$$\boldsymbol{V} = \boldsymbol{V}^{base} + V^{twap}\boldsymbol{a}, \qquad (1)$$
$$\text{s.t. } \boldsymbol{a}_l \leq \boldsymbol{a} \leq \boldsymbol{a}_u,$$

where $V^{twap}$ is the volume per volume calculated from the TWAP-like strategy, $\boldsymbol{V}^{base} = [V^{twap}, 0, 0]$ is the order volume vector for the baseline strategy, $\boldsymbol{a}$ is the agent's action vector, $\boldsymbol{a}_l = [-1, 0, 0], \boldsymbol{a}_u = [3, 1, 1]$ are the lower and upper bounds of the agent's actions, respectively, $\boldsymbol{v}$ is the actual order volume vector. It can be seen from (1) that the RL agent not only solves the optimal execution problem, but also solves the optimal placement problem as it has control of orders over different LOB levels. Such simple action implementation has two main advantages. First, the agent does not need to learn the baseline TWAP policy (requires a neural network to generate the same output values regardless of the inputs) which may be difficult to realize. Second, radical orders with large volumes can be easily avoided by constraining the action's bounds. This ensures the inventory risk and execution risk are at a reasonable level.

Besides the aforementioned action formulation, we also tried different action formulations that allow the RL agent to directly output order volumes. The first attempt directly used the scaled action values as the order volume $\boldsymbol{V} = V_0^r \boldsymbol{a}$, where $V_0^r$ is the total volume of the execution task (remaining quantity at time step 0). This action formulation requires variable $\boldsymbol{a}_u$ to ensure the feasibility of the agent's action, as the order volume (summed over all order levels) cannot exceed the remaining quantity from the execution task. Although such variable upper bound can be implemented, it requires the network to learn much finer control in the small action range. In practice, we observed that the agent's behaviour tends to diverge due to radical orders at the beginning of the trading period, or converge to sub-optimal strategies with action continuously hitting the upper bound. The second attempt is to generate linear/nonlinear mapping such that the full action range spans across the remaining volume. However, these implementations severely degrade the learning efficiency, and the RL agent rarely converges to good performance. We suspect that the additional mapping between the action and the order volume introduces confusion to the agent as the same action values mean different order volumes depending on the remaining quantity.

### 2.3 Reward Formulation

The objective for optimal execution in a BUY[4] setting is to minimize the expected total execution cost using the following equation [2, 20]:

$$\min_{V_1^e,...,V_T^e} E\left[\sum_{t=1}^{T} P_t^e V_t^e\right], \text{ s.t.} \sum_{t=1}^{T} V_t^e = V_0^r, \quad (2)$$

where $P^e$ and $V^e$ are the execution price and volume for individual orders, $T$ is the maximum time periods allowed for the trading period, $V_0^r$ is the total volume. However, this objective is difficult to evaluate as it requires the whole trading history upon completing the execution task. To improve the learning efficiency of the RL agent, we propose our reward formulation based on the idea of imitation learning where a teacher agent (i.e., a TWAP agent in this study) is added to the same environment to provide intermediate evaluations of the agent's performance. The proposed reward function consists of two terms, including imitation reward and competitive reward. To explain the mathematical formulation, we first define the execution cost $\mathcal{C}$ and the execution volume $\mathcal{V}$ at time step $k$ with a positive temporal window size $j$ (a hyperparameter) as follows:

$$\mathcal{C}_k = \sum_{t=k-j+1}^{k} P_t^e V_t^e, \quad \mathcal{V}_k = \sum_{t=k-j+1}^{k} V_t^e, \quad \text{s.t.} 1 \leq j \leq k \leq T. \quad (3)$$

The imitation reward encourages the agent to mimic the teacher's strategy by matching the teacher's execution volumes as shown below:

$$R_{mimic} = -f(\mathcal{V}_k^{RL}, \mathcal{V}_k^{base}), \quad (4)$$

where $\mathcal{V}_k^{RL}, \mathcal{V}_k^{base}$ are the execution volumes at time $k$ for the RL agent and the teacher, respectively, $f(\cdot)$ is some distance measure (L2 norm in our implementation). During the initial learning phase, this imitation reward allows the agent to transfer the teacher's market order based strategy to its own limit order based strategy. The imitated strategy also improves the later learning efficiency by avoiding reward divergence caused by bad strategy initialization.

The second reward term encourages the agent to construct its own strategy by competing and beating the teacher. This competitive reward is calculated based on the teacher's execution cost $\mathcal{C}_k^{base}$ and the agent's execution cost $\mathcal{C}_k^{RL}$. However, during the trading period, $\mathcal{V}_k^{base}$ likely differs from $\mathcal{V}_k^{RL}$, which makes the execution cost comparison unfair. To address this issue, we perform hypothetical volume matching to calculate the potential gain/loss at time $k$. Here we skip the subscript $k$ for clarity purposes.

$$R_{comp} = \hat{\mathcal{C}}^{base} - \hat{\mathcal{C}}^{RL} \quad (5)$$

$$= \left[\mathcal{C}^{base} + \left(\mathcal{V}^{RL} - \mathcal{V}^{base}\right) P\right] - \mathcal{C}^{RL}, \quad (6)$$

$$= \mathcal{C}^{base} + \Delta \mathcal{V} P - \mathcal{C}^{RL}, \quad (7)$$

where $\hat{\mathcal{C}}$ is the hypothetical cost after volume matching, $P$ is the level-one price at time $k$. For simplicity, the feasibility of fulfilling $\Delta \mathcal{V}$ based on actual volumes available on the LOB is not checked. The final reward function combines the imitation reward and the competitive reward as below:

$$R = R_{comp} + \alpha R_{mimic}, \quad (8)$$

where $\alpha$ is a hyperparameter that controls the balance between imitation and generation of agent's own strategies. Both $j$ in (3) and $\alpha$ in (8) are tuned during the hyperparameter optimization explained in section 4.

---

[4]Buy or sell (liquidation) settings do not affect the fundamental or objective of the execution problem.

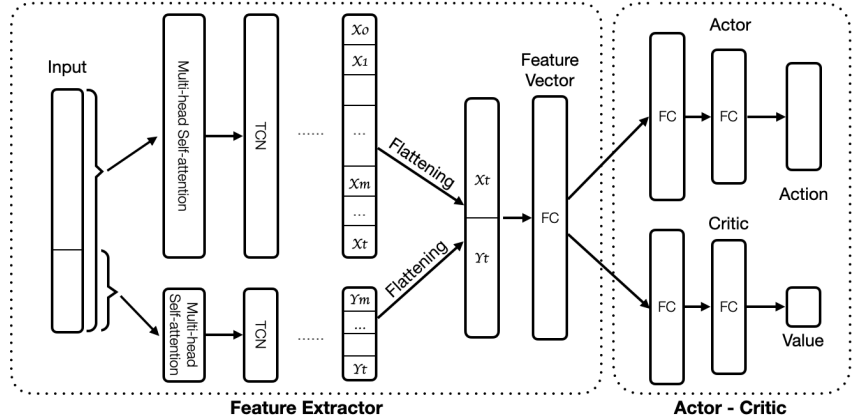

Figure 2: The architecture of Dual-Window Denoise PPO network; the feature extractor condenses the temporal market observations and outputs feature vectors to the Actor-Critic networks. Two parallel networks in the feature extractor are responsible for processing long-term (top) and short-term (bottom) market information. FC represents fully connected networks.

## 3 Dual-Window Denoise PPO

The optimal execution problem involves continuous state observations and actions. We choose the Proximal Policy Optimization (PPO) [12] algorithm as the stem of our agent implementation due to its stable convergence. To adapt to the specific needs of learning in noisy financial markets, we developed the Dual-Window Denoise PPO with the agent's network architecture presented in Fig. 2. It comprises a common feature extractor and a pair of actor-critic networks. The feature extractor condenses the high-frequency market observations into a feature vector and then passes it to the actor-critic networks for decision making and learning. Here, we explain on our design rationale for the feature extractor in the proposed Dual-Window Denoise PPO architecture.

Financial markets are often assumed Markovian for simplicity of analysis / mathematical formulation [7] but may not be the case in reality. It is also unlikely that our feature selection can perfectly suffice the Markov decision process, therefore, we assume a non-Markovian process and allow temporal information as observation and let the agent recover potential hidden valuable information. To accomplish this, we configure the market simulation to provide minute-level market information and use the temporal convolutional network (TCN) [35] to efficiently learn the latent features from the market history. As high-frequency market data can be noisy, unfiltered observation can degrade RL learning efficiency at the early stage [36]. To combat this, we utilize the multi-head self-attention layers [37] as denoising filters (inspired by [38] at the beginning of the network. In addition, to effectively capture both the long-term trends and short-term market behaviours, we added two parallel feature extractors shown in Fig. 2 with different receptive fields. Their outputs are then concatenated and mapped to the feature vector. We ensure the same length of the long-term and the short-term vectors prior to concatenation.

## 4 Multi-agent Training

Prior to training, five hyperparameters including three PPO-related terms (epochs for batch update $n_e$, trajectory length $l_t$, discount factor $\gamma$) and two reward-related terms ($j$ in (3), $\alpha$ in (8)) were tuned using grid search, where three values for each hyperparameter were sampled using exponential intervals. Due to computational constraints, we separated the tuning process into two phases, where the first phase optimized the PPO-related terms (total of 27 experiments), and the second phase optimized the reward-related terms (total of 9 experiments). During hyperparameter search, LOB data from only one ticker (IBM) is used to avoid overfitting. The final hyperparameters ($n_e$=10, $l_t$=64, $\gamma$=0.99, $j$=64, $\alpha$=0.01) are then used for all later experiments.

NASDAQ nano-second level message data from five tickers are used to replay the LOB and train the RL agent. Each ticker has consecutive 19 days of data. We used the first 12 days for training and the last 7 days for testing[5] (testing dates are consistent across all tickers). Each training experiment only uses LOB data from one ticker so that inter-ticker and intra-ticker generalization can be both evaluated. The reported results for both inter-ticker and intra-ticker performance are based on the testing dates that do not overlap with the training dates.

Table 2: Agent performance with respect to the TWAP strategy ($\Delta C$ and median are in ‰).

| Ticker | $\Delta C$ (mean ± SD) | median | GLR | $\mathbb{P}(\Delta C > 0)$ |
|---|---|---|---|---|
| INTC | 2.94 ± 3.15 | 3.20 | 2.18 | 0.57 |
| YHOO | 2.04 ± 3.99 | 2.23 | 2.58 | 0.71 |
| IBM  | 0.30 ± 1.63 | 0.39 | 2.18 | 0.57 |
| CSCO | 1.20 ± 1.63 | 0.93 | 4.00 | 0.71 |
| MSFT | 0.71 ± 0.76 | 0.47 | 5.66 | 0.86 |

Table 3: Agent generalization across tickers measured using $\Delta C$ (mean ± SD in ‰).

|  |  | tickers for testing | | | | |
|---|---|---|---|---|---|---|
|  |  | INTC | YHOO | IBM | CSCO | MSFT |
| tickers for training | INTC | N/A | 2.06 ± 4.07 | −0.24 ± 2.34 | 1.35 ± 2.01 | 0.77 ± 0.86 |
|  | YHOO | 2.86 ± 3.06 | N/A | −0.22 ± 2.38 | 1.29 ± 1.91 | 0.87 ± 1.02 |
|  | IBM | 2.41 ± 2.61 | 2.01 ± 3.78 | N/A | 1.05 ± 1.71 | 0.24 ± 0.35 |
|  | CSCO | 2.42 ± 2.68 | 2.03 ± 2.80 | −0.22 ± 2.10 | N/A | 0.36 ± 0.51 |
|  | MSFT | 2.75 ± 2.96 | 2.02 ± 3.94 | −0.23 ± 2.22 | 1.20 ± 1.90 | N/A |

Table 2 summarizes the intra-ticker performance of the agents based on three metrics. The relative performance $\Delta C$ with respect to the competitive TWAP agent is calculated as $\Delta C = -\frac{C_{RL} - C_{twap}}{C_{twap}}$, where $C$ represents the cost of the execution task over the entire episode (day). The gain-loss ratio (GLR) inspired by [13] is calculated using GLR = $\frac{\mathbb{E}[\Delta C | \Delta C < 0]}{\mathbb{E}[-\Delta C | \Delta C > 0]}$. The gain probability $\mathbb{P}(\Delta C > 0)$ calculates the ratio between the number of RL-outperformed days and the total number of days tested.

Intra-ticker performance suggested that the trained agents were capable of generalizing over unseen dates with great performance beating the TWAP agent for all tickers (indicated by the positive $\Delta C$ mean and median). GLR values were greater than 1 indicating that the average gain from the RL-outperformed dates was larger than the average loss from the RL-underperformed dates. The probabilities of RL outperforming are all greater than 0.5 showing more outperforming dates than the underperforming ones. Despite the great performances, performance differences were observed between tickers. These differences might be caused by different behaviours of market participants and the agent's convergence to different local optima.

The inter-ticker performance of the trained agents is summarized in Table 3. Each trained agent was tested on four untrained tickers and $\Delta C$ (mean±SD) is provided. All five agents were able to generalize over most of the unseen tickers indicated by positive mean values of $\Delta C$ in Table 3. However, the performance on IBM did not show a lead compared to the TWAP strategy. This aligns with the intra-ticker testing in Table 2 where the $\Delta C$ for IBM is the least significant among all five tickers. We suspect that there might be some market-related challenges (e.g., insufficient liquidity and high volatility) contributing to learning difficulties. Regardless, the performance degradation for IBM in Table 3 is small and can be easily overcome by fine-tuning the models using IBM LOB data.

We also performed four ablation studies to validate the superiority of our RL problem formulation. The first ablation study used a simplified policy network that only had 3-layer-64-neuron feedforward networks. The second ablation study used a simplified feature set where only raw LOB data and task-related features were included. Third, the agent's action was simplified to allow only market order with the action limit of $[-1, 3]$ (i.e., 0% to 400% of TWAP market order quantity). Last, we set $\alpha$ in (8) to 0 so that the imitation reward was turned off. The aggregated results for all tickers were summarized in Table 4. It was clear that the agent's performance degraded after removing either of the four components, particularly when agent's ability for placement is restricted. In the simplified action case, the large average GLR was caused by a single large GLR value, and it is no longer meaningful given its low outperforming probability.

---
[5] All data available to us has been utilized in the study. We are open to any data sponsorship to carry out large-scale experiments in the future.

Table 4: Agent ablation study with average $\Delta C$ across all tickers ($\Delta C$ and median are in ‰).

| Conditions | $\Delta C$ (mean $\pm$ SD) | Median | GLR | $\mathbb{P}(\Delta C>0)$ |
|---|---|---|---|---|
| No ablation | **1.44$\pm$2.52** | **1.44** | 3.89 | **0.71** |
| Simplified network (MLP only) | $0.91 \pm 2.16$ | 0.99 | 3.23 | 0.60 |
| Simplified features (Raw LOB only) | $0.99 \pm 2.24$ | 1.14 | 3.34 | 0.60 |
| Restricted action (market order only) | $0.24 \pm 1.37$ | 0.40 | **4.85** | 0.49 |
| Simplified reward (no imitation) | $1.11 \pm 2.08$ | 1.22 | 3.60 | 0.66 |

## 5 Conclusions and Future Work

In this paper, we built an OpenAI Gym compatible historical LOB simulator where price impact and transaction cost can be correctly assessed. We combined this realistic market simulator with a rich set of market features, a flexible agent action formulation, an imitation-based reward scheme, and the Dual-window Denoise PPO in synergy, leading to our novel RL framework for solving the optimal execution and placement problems. As a result, our model-free execution agents learned effective strategies from the noisy environment and consistently beat the baseline TWAP benchmark. Our strategies also demonstrated strong generalization ability when they were evaluated on out-of-sample dates and securities.

In future studies, a richer set of historical data will be used to further validate the performance of our RL agents on execution tasks. Interpretability of the RL agent's strategies and the feature importance will be investigated as well. We also look forward to extending our proposed simulator and methodologies to other trading problems with more complex objectives and strategies, such as high-frequency market making and arbitrage.